\documentclass[10pt,journal,compsoc]{IEEEtran}
\ifCLASSOPTIONcompsoc
  \usepackage[nocompress]{cite}
\else
  \usepackage{cite}
\fi
\ifCLASSINFOpdf
  \usepackage[pdftex]{graphicx}
\else
\fi
\usepackage{amsmath}
\usepackage{amsfonts}
\usepackage{bm}
\DeclareMathOperator*{\argmax}{arg\,max}

\newtheorem{theorem}{Theorem}[section]
\newtheorem{Proposition}[theorem]{Proposition}

\usepackage{comment}

\usepackage{algorithmic}
\newcommand{\algcaption}[1]{%
  \par\noindent
  \textbf{Algorithm:}
  #1\par
}

\usepackage{array}
\usepackage{booktabs}       % professional-quality tables
\usepackage{multirow}

\usepackage{url}
\begin{document}
\title{Functional Classwise Principal Component Analysis: A Novel Classification Framework}

\author{Avishek~Chatterjee,~\IEEEmembership{}
        Satyaki~Mazumder,~\IEEEmembership{}
        Koel Das~\IEEEmembership{}% <-this % stops a space
\IEEEcompsocitemizethanks{\IEEEcompsocthanksitem A. Chatterjee, S. Mazumder and K. Das are with the Department
of Mathematics and Statistics, IISER Kolkata, Mohanpur, West Bengal, India, 741246.\protect\\
E-mail: koel.das@iiserkol.ac.in
%\IEEEcompsocthanksitem A. Chatterjee and S. Mazumder are with with the Department
%of Mathematics and Statistics, IISER Kolkata, Mohanpur, West Bengal, India, 741246.
}% <-this % stops an unwanted space
\thanks{ }}

\IEEEtitleabstractindextext{%
\begin{abstract}
In recent times, functional data analysis (FDA) has been successfully applied in the field of  high dimensional data classification. In this paper, we present a novel classification framework using functional data and classwise Principal Component Analysis (PCA). Our proposed method can be used in high dimensional time series data which typically suffers from small sample size problem. Our method extracts a piece wise linear functional feature space and is particularly suitable for hard classification problems. The proposed framework converts time series data into functional data and uses classwise functional PCA for feature extraction followed by classification using a Bayesian linear classifier. We demonstrate the efficacy of our proposed method by applying it to both synthetic data sets and real time series data from diverse fields including but not limited to neuroscience, food science, medical sciences and chemometrics. 
\end{abstract}

}

\maketitle

\IEEEdisplaynontitleabstractindextext

% \fi
\IEEEpeerreviewmaketitle

\IEEEraisesectionheading{\section{Introduction}\label{sec:introduction}}

\IEEEPARstart{T}{he} advent of modern technologies has given rise to capability of generating both continuous and discrete data at an unprecedented scale. Analyzing such a large volume of data poses a veritable challenge in Big data analytics.  In recent times, functional data analysis (FDA) has emerged as a successful alternative to  traditional methods in big data analysis \cite{hadjipantelis2018functional} %(ref) %(https://link.springer.com/chapter/10.1007/978-3-319-18284-1_18)
and has become increasingly popular choice of data analysis techniques in various fields including biology, economics, public health, environmental studies and geology \cite{ullah2013applications, hall2006properties,ramsay2007applied,ullah2010functional}. Functional data analysis \cite{ramsay2004functional} is a statistical framework where data are expressed in terms of a continuous functional entity of infinite dimension and discrete data samples are then treated as finite realizations from the continuous underlying stochastic process. The underlying function is initially estimated, and subsequent analyses are then performed on the collection of functional data by using tools from the field of functional data analysis. 
 Functional data provides a rich source of information for data interpretation and analysis \cite{wang2016functional}. One of the main advantages of smooth functional data is that it provides accurate estimates of parameters, to be used in subsequent analysis, that do not suffer from shortcomings of model specifications. FDA avoids errors occurring due to  presence of noise in the data by using smooth functions. It can be applied to data having irregularly placed time points and correlated data points and can be used to infer additional information present not only in its functional form but also at the level of derivatives. Small sample size problem which often acts as a major hindrance in high dimensional data analysis does not pose a challenge in FDA as the data is converted into a functional form. The advantages of FDA makes it an ideal contender for its application in feature extraction and classification of time-series data.
 
 Predictive data analysis is one of the major application areas in FDA. Predictive models for functional data generally adapts the idea of standard linear models and the functional principal component analysis (FPCA) is commonly used for better estimation of the parameter functions of such models \cite{aguilera2000principal,escabias2004principal}. Functional linear models (FLM) \cite{ramsay2009introduction,kadri2011multiple}  frequently appear in FDA methods and in FLM, the inner product between functional covariate and an coefficient function is  used to estimate the effect of functional covariates on response. But in many situations only a few out of many functional covariates are actually useful in predicting the response \cite{gertheiss2013variable} or multicollinearity of the covariates makes the model inconsistent in model estimation \cite{preda2007pls}. In such situations, a reduced number of functional principal components (FPCs) of the original variables can be used as covariates of the model to get a better estimation of the model parameters \cite{escabias2004principal}. As an alternate to functional-PCs, Preda et al.~\cite{preda2007pls} used Partial Least Squares (PLS) components to perform linear discriminant analysis (LDA) for functional predictors \cite{preda2007pls} since LDA can not be directly applied to functional data because of its infinite dimension. Apart from other regularization methods \cite{friedman1989regularized,hastie1995penalized}, James and Hastie developed a functional-LDA model \cite{james2001functional} for irregularly sampled functional data using cubic spline functions that represents the data and the corresponding coefficients. 
 
 A common feature to most FDA based classification techniques is that
they typically project the data in a single (linear) functional subspace during data reduction stage and classification is then performed on the reduced functional feature space. However, projecting the data to multiple functional feature space can capture the underlying manifold of the input data and may result in better classification performance. 
In this paper, we present a novel supervised classification framework based on functional data analysis and local Principal Component Analysis that can be used to classify successfully time-series data from various fields. The proposed  method Functional Classwise Principal Component Analysis (FCPCA) utilizes the strength of FPCA as a functional dimensionality reduction technique and improves on FPCA by preserving class-specific information by applying the concept of classwise projection of functional data giving rise to a piecewise linear functional feature space. Classification is subsequently carried on the reduced functional feature space using Bayesian Linear Classifier.
Although we have chosen to use a linear classifier in the current study, FCPCA provides a framework whereupon after projection of data onto functional subspace, any classifier of choice can potentially be used for classification. We evaluate our proposed method using several benchmarked times series datasets and novel simulated datasets and compare the performance with several state-of-the-art machine learning classifiers along with standard functional data classification techniques. The results demonstrate the efficacy of our proposed method both in terms of performance improvement and computational time.
 
 The remaining part of this paper is organized as follows: Section 2 describes the preliminaries required for our proposed method, FCPCA, Section 3 explains FCPCA in details, Section 4 shows experimental results based on synthetic data and real data including classification benchmarks and concluding remarks are provided on Section 5.

 \section{Preliminaries}
\label{related_work}
 Data in many real-world fields, such as medicine, chemometrics, neuroscience, economics and many others \cite{ramsay2007applied} can be naturally represented as functions. In this work, we used the functional representation of data.
 Let $X_1,X_2,\ldots,X_N$ be the functional observations of a stochastic process $\{X(t):t \in T \}$, where $ T $ is a bounded interval and $ N $ is the number of functional observations. From now on, we will assume the following: The stochastic process is of second order, continuous in quadratic mean; $X_i(t), t\in T$, for $i=1, \ldots, N,$ are squared integrable and the sample paths belongs to the Hilbert space $L^2(T)$ with the usual inner product defined as
 \begin{equation}
    \langle F,G \rangle = \int_{T} F(t)G(t)dt \;\;\; \forall F,G \in L^2(T).
\end{equation}
In practice, no machinery, how sophisticated it is, can observe a set of functions continuously in time. Instead we get observations of such functions in a set of discrete time points $\{t_{i_0}, t_{i_1},\ldots,t_{i_m}:i=1,2,..,N\}$. In this article, we only consider regularly sampled curves, i.e., for each $i\in \{1, \ldots, N\}$, the number of time points are the same, say m, and for each $k\in \{1, \ldots,m\}$, $t_{1_k} = \ldots = t_{N_k}$.
\par 
\noindent
The first step in FDA is to transform the raw discrete data into smooth functional data. To achieve this linear combination of basis function can be used. The most common choices of such basis function systems are Fourier and
B-splines \cite{ramsay2013functional}. Methods of smoothing or interpolation can be used to obtain the functional form of the sample curves by approximating the basis coefficients \cite{escabias2004principal}.
\par 
In our proposed method Functional principal component analysis (FPCA) is used locally on each class as a natural choice of dimension reduction technique for functional data classification. Two key components of our proposed method, FPCA and  Gram-Schmidt orthonormalization for functional data are described briefly.

\subsection{Functional principal component analysis}
\label{Section: FPCA}

FPCA is one of the most widely used techniques in FDA and similar to Principal component Analysis (PCA) on multivariate data, FPCA is used to reduce the dimension of 
infinitely dimensional functional data to a small finite dimension in an optimal way by finding an orthonormal basis that captures the maximum variability of the data. We describe the FPCA method in brief next.

%We focus empirical aspects while defining FPCA.\\
Let $ X_i $s denote the functional observations as before. The sample mean and the covariance functions of the sample curves can be estimated, respectively, as
\begin{equation}
    \hat{\mu}(t) = \frac{1}{N} \sum _{i=1}^N X_i(t) \mbox{ and }
\end{equation}
\begin{equation}
    \hat{C}(t,s) = \frac{1}{N-1} \sum _{i=1}^N (X_i(t)- \hat{\mu}(t))(X_i(s)- \hat{\mu}(s)).
\end{equation}

Under the assumption that $\hat{\mu}(t) =0 \;\; \forall t \in T$, the functional PCs of $X_1,X_2,\ldots,X_N$ are defined as follows:
$$\xi_{ij}=\int_{T} X_i(t)f_j(t)dt \;\;\;\; i=1,2,\ldots,N,$$
where the weight functions $f_j(t), \, j=1,2,\ldots,N-1$ are nothing but the eigenfunctions of the covariance operator corresponding to the positive eigenvalues $\lambda_j, \, j=1,2, \ldots ,N-1$. 

\subsection{Gram-Schmidt orthonormalization for functional data}
\label{GS orthogonalization}

Gram-Schmidt orthogonalization is an important and fundamental concept in linear algebra. The main idea of this technique is to transform a set of non-orthogonal   linearly independent functions to an orthogonal basis over an arbitrary interval \cite{huang2001recursive}. Gram-Schmidt orthogonalization procedure has numerous applications in various areas \cite{bjorck1967solving,korenberg1988orthogonal,ge1998iterative}. The technique can easily be incorporated for functional data. Liu et al. \cite{liu2018functional} used the Gram-Schmidt orthogonalization process as a variable (functional) selection tool to perform multiple functional linear regression. We provide a short description of Gram-Schmidt transformation on functional data below.
\par 
\noindent
Let $F(t)=\{f_1(t),\ldots,f_n(t)\}$ be  a set of non-orthogonal  linearly independent functions defined over an interval $T$. Now the transformation aims to constructs a sequence of orthogonal functions $G(t)=\{g_1(t),\ldots,g_n(t)\}$ from $F(t)$ as follows:
\begin{equation}\label{eq1}
\begin{split}
g_1(t) & = f_1(t) \\
g_k(t)& = f_k(t)-\sum _{j=1}^{k-1}\frac{\langle f_k(t),g_j(t) \rangle}{\langle g_j(t), g_j(t) \rangle}g_j(t),\, k=2, \ldots,n,\\
 %g_2(t)& = f_2(t)-\frac{\langle f_2(t),g_1(t) \rangle}{\langle g_1(t), g_1(t) \rangle}g_1(t)\\
% &\vdots\\
% g_n(t)& = f_n(t)-\sum _{j=1}^{n-1}\frac{\langle f_n(t),g_j(t) \rangle}{\langle g_j(t), g_j(t) \rangle}g_j(t)\\
\end{split}
\end{equation}
where  $\mbox{span}\{f_1(t),\ldots,f_n(t)\}=\mbox{span}\{g_1(t),\ldots,g_n(t)\}$. 
Further, orthonormal set $G^{\prime}(t)$ can be formed by normalizing each of $g_k(t)$ by $||g_k||=\sqrt{\langle g_k(t),g_k(t) \rangle}$, i.e.,
\begin{equation} \label{eq2: orthonormal}
    G^{\prime}(t)=\left\{\frac{g_1(t)}{||g_1||},\frac{g_2(t)}{||g_2||},\ldots,\frac{g_n(t)}{||g_n||} \right\}.
\end{equation}
The following proposition relating Gram-Schmidt orthogonalization process has been used in the functional feature extraction part of our proposed method.
%\textbf{Proposition:} 
\begin{Proposition}
\label{propostion1}
Let $A$ be a finite subset of linearly independent functions in the Hilbert space $L^2(T)$ and $B \subseteq A$ is an orthonormal set. Then $B$ remains unchanged after Gram-Schmidt orthonormalization process on $A$.
\end{Proposition}

\textbf{Proof:}
Let, $A=\{f_1,f_2,\ldots, f_n \}$, where $f_i \in L^2(T)$ and without loss of generality the subset $B$ of $A$ is chosen as $B=\{f_1,f_2,\ldots,f_p \}, $ where $p \leq n$. Since, $B$ is orthonormal, for $i,j \in \{1, \ldots, p\}$, $\langle f_i,f_j \rangle=0$ when $ i \neq j$, and $||f_i||=1$. 
Applying Gram-Schmidt orthonormalization process on $A$ and using the fact that $\langle f_i,f_j \rangle=0, \, \forall i\neq j \in \{1, \ldots, p\}$, yields an orthogonal set of function $G=\{g_1,g_2,\ldots,g_n \}$ such that,

\begin{equation*}\label{eq2}
\begin{split}
g_1(t) & = f_1(t) \\
 g_2(t)& = f_2(t)-\frac{\langle f_2(t),g_1(t) \rangle}{\langle g_1(t), g_1(t) \rangle}g_1(t)=f_2(t) 
 %\;\;\;\mbox{since,}\; g_1(t)=f_1(t)\; \mbox{and}\; \langle f_1(t),f_2(t) \rangle=0
 \\
 &\vdots\\
 g_p(t)& = f_p(t) 
 %\;\;\;\mbox{since,} \; g_i(t)=f_i(t)\;\forall i=1,2,\ldots,p-1 \;\mbox{and} \;\langle f_i,f_j \rangle=0 \;\forall i,j \; \mbox{and}\; i \neq j
 \\
 g_{p+1}(t)& = f_{p+1}(t)-\sum _{j=1}^{p}\frac{\langle f_{p+1}(t),g_j(t) \rangle}{\langle g_j(t), g_j(t) \rangle}g_j(t)\\
 &\vdots\\
 g_n(t)& = f_n(t)-\sum _{j=1}^{n-1}\frac{\langle f_n(t),g_j(t) \rangle}{\langle g_j(t), g_j(t) \rangle}g_j(t)\\
\end{split}
\end{equation*}
Further, orthonormal set $G^{\prime}$ can be formed by normalizing each of $g_i(t)$ by $||g_i||$. Hence, $G^{\prime}$ can be written as follows.
\begin{equation*} \label{eq_proof}
  G^{\prime}=  B \cup \{\frac{g_{p+1}(t)}{||g_{p+1}||},\frac{g_{p+2}(t)}{||g_{p+2}||},\ldots,\frac{g_n(t)}{||g_{n}||} \}.
\end{equation*}
Hence, $B$ is unchanged after the  Gram-Schmidt orthonormalization process. \hfill $\square$

\section{Functional Classwise principal component analysis}
\label{S:4}
 FCPCA exploits the dimension reduction property of FPCA  and performs classification for functional data by preserving the class-specific information of the data set. The key idea of the method is based on class-wise FPCA which helps to identify and get rid of the non-informative subspaces of the data using  a piece-wise linear mapping to a lower-dimensional space. 
Similar concept of employing PCA locally has been found to enhance performance in both supervised \cite{das2009efficient,min2004locality} and unsupervised learning \cite{liu2003improved}  techniques.
FCPCA performs  multi-class classification problems while producing $c$ number of subspaces $\{S_1,S_2,\ldots,S_c \}$ (for a $c$-class classification problem) and then selecting the one which gives the best classification accuracy. The details of the algorithm will be described for regularly sampled curves for multiple classes, where $c \geq 2$.

\begin{figure*}[!t]
\centering
\includegraphics[width=5 in]{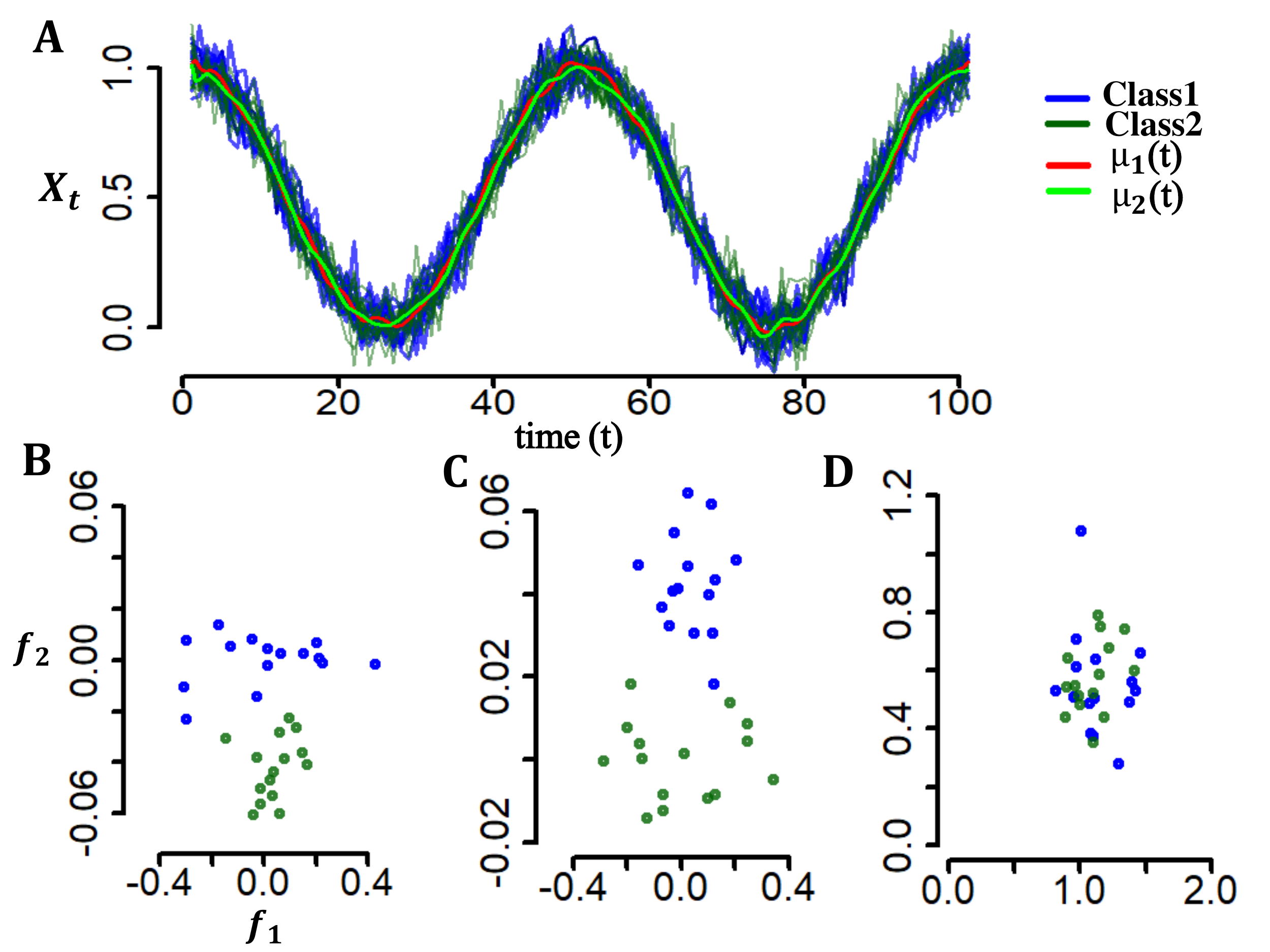}
\caption{ (A) Simulated Functional data: Two classes ($\omega_i \; i=1,2$) each having 15 samples. Corresponding mean functions ($\mu_i(t)
\;i=1,2$) are also plotted. (B) Projection of the data to the 2 D augmented functional principal component subspace $S_1$ of $\omega_1$, where $f_1$ and $f_2$ represent the first PC and orthonormalised mean difference function ($\mu(t)-\mu_1(t)$) respectively. (C)  Projection of the data to the 2 D augmented functional principal component subspace $S_2$ of $\omega_2$. (D) Projection of the data to the global functional principal component subspace. }
\label{example3}
\end{figure*}

To motivate the usefulness of FCPCA in classification over FPCA, we present a toy example here. We consider the data of first class and second class being generated from the models
\begin{align*}
    y(t)=\cos^2(2\pi t)+\epsilon_t \text{ and } y(t)=\cos^2(2\pi (t-a))+\epsilon_t,
\end{align*}
%\begin{align*}
%    y(t)=\cos^2(2\pi (t-a))+\epsilon_t,
%\end{align*}
respectively,
where $t\in [0,1]$, $a=1$ and $\epsilon_t\sim N(0,0.07),\, \forall t$ .  The functional data and the mean functions are plotted in Figure \ref{example3} (A). In Figure~\ref{example3} (B) and (C), the coefficients of the projected functions in two spaces, obtained using FCPCA, are depicted. The coefficients of projected functions using FPCA is shown in Figure~\ref{example3} (D). The figures~\ref{example3} (B), (C) and (D) clearly show that the class separability when using FCPCA is superior  when compared to FPCA.

\subsection{Functional feature extraction}
%Let $\omega_i$  $ (i=1,2,\ldots,c) $ denotes $c$ number of classes with sample means $\hat{\mu}_i(t)$  and covariance functions $\hat{C}_i$ and each class contains $n_i$ number of functional observations (i.e., $X_{1,i},X_{2,i},\ldots,X_{n_i,i} \in \omega_i$). 
Let $\omega_1, \ldots, \omega_c,$  %$i=1,2,\ldots,c$, 
denote $c$ number of classes. Assume that $\omega_i$ contains $n_i$ number of functional observations, $X_{1,i},X_{2,i},\ldots,X_{n_i,i}$. We denote the sample mean and the sample covariance function, calculated from $X_{1,i},X_{2,i},\ldots,X_{n_i,i}$, by $\hat{\mu}_i(\cdot)$ and $\hat{C}_i(\cdot,\cdot)$, respectively, corresponding to the class $\omega_i$, for $i=1, \ldots, c$. 
Now our aim is to classify an unknown (test) sample curve $X^\ast$. To achieve this, first we project $X^\ast$ in  $c$ number of subspaces $\{S_1,S_2,\ldots,S_c \}$ using the following mappings.
\begin{equation}\label{projection1}
    X_{(i)}^\ast= \sum_{k=1}^{q_i+c-1} \langle X^\ast-\hat{\mu_i}, f_{ki} \rangle f_{ki},  \;\;\;\;\;\; i=1,2,\ldots,c, 
\end{equation}
where the elements of the set $F_i=\{ f_{1i},f_{2i},\ldots,f_{q_ii} \}$ are the $q_i$ ($q_i$ to be chosen)  empirical  functional PCs of the class $\omega_i$. Additionally, to get rid  of overlapping class projections, the set of functions $G= \{\hat{\mu}(t)-\hat{\mu_1}(t), \hat{\mu}(t)-\hat{\mu_2}(t),\ldots,\hat{\mu}(t)-\hat{\mu}_{c-1}(t) \}$ is augmented along with the functional PCs, where $\hat{\mu}(t)$ denotes the grand mean. To keep the direction of the projections orthogonal after augmenting with $G$, the  set of functions ($\{F_i, G\}$)  are orthonormalized using Gram-Schmidt Orthonormalization process (see equations (\ref{eq1}) and (\ref{eq2: orthonormal})). After applying Gram-Schmidt orthonormalization technique, if any function turns out to be identically zero, then it is discarded from the orthonormal set. In that case, the cardinality of the fully orthonormalized set will be less than $q_i+c-1$. However, for simplicity, let us assume that the fully orthonormal set will contain $q_i+c-1$ orthonormal functions generating subspace $S_i$. 
%To continue the discussion, we present here a proposition which ensures that the functional PCs which are an orthonormal basis of the subspace $S_i$, remain unchanged in the Gram-Schmidt Orthonormalization process. Thus only remaining $(c-1)$ augmented functions need to be modified while performing Gram-Schmidt Orthonormalization process.
Since the functional PCs form an orthonormal basis of the subspace $S_i$, they will remain unchanged in the Gram-Schmidt Orthonormalization process (see \ref{propostion1} for proof)
%(proof is immediate from equation (\ref{eq1})) 
and the remaining $(c-1)$ augmented functions are modified and renamed as
$G_i^\ast=\{ f_{(q_{i}+1)i}, f_{(q_{i}+2)i},\ldots,f_{(q_{i}+c-1)i}\}$. 
Using similar projections (as in equation(\ref{projection1})) all the training sample curves $X$ (total number $N=\sum\limits_{i=1}^{c}n_i$) are represented in $c$ number of subspaces, which yields $c$ number of coefficient matrices, viz. $\{Z_i|i=1,2,\ldots c\}$ as follows: 
\begin{equation}
\label{coefficient matrix}
\begin{split}
     X_{(i)} & = \sum_{k=1}^{q_i+c-1} \langle X-\hat{\mu_i}, f_{ki} \rangle f_{ki},  \;\;\;\;\;\; i=1,2,\ldots,c  \\
  {} & = Z_i[F_i | G_i^\ast], \\
\end{split}
\end{equation}
where $[F_i | G_i^\ast]$ is the augmented set of functions for the subspace $S_i$. 
Functional feature extraction part of our proposed method is implemented through the following algorithm.
%\begin{algorithm}
%\caption{Functional Feature Extraction (Training)}

\algcaption{Functional Feature Extraction (Training)}

\begin{algorithmic}[1]

    \STATE{\textbf{begin}} compute $\mu (t)$

       \FOR{$i \gets 1$ to $c$}     
       
        \STATE{\textbf{classwise FPCA:}} compute $F_i$
        \STATE{\textbf{augment:}}  $E_i=[F_i | G_i]$
        \STATE{\textbf{apply}  Gram-Schmidt orthonormalization process on $E_i$ and get $E_i^\ast=[F_i | G_i^\ast]$ }
        \IF{there are few identically zero columns }
                \STATE{discard the identically zero columns from $E_i^\ast$}
        \ENDIF
         \STATE{\textbf{projections:}} compute  $Z_i= \langle X-\hat{\mu_i}, e \rangle e$, $e \in E_i^\ast$
        
        \ENDFOR
        \STATE{\textbf{return:}} $E_1^\ast,E_2^\ast,\ldots,E_c^\ast$, $Z_1,Z_2,\ldots,Z_c$
        \STATE{\textbf{end}}

\end{algorithmic}
%\end{algorithm}

\subsection{Classification}
Given an unknown functional observation $X^\ast$, we obtain the coefficient vector in the $i$th subspace as $(\langle X^\ast-\hat{\mu_i}, f_{1i} \rangle, \ldots, \langle X^\ast-\hat{\mu_i}, f_{q_i+c-1,i} \rangle)'$, which we denote as $\mathbf{v}_i$. Then we calculate the posterior probabilities of the class $\omega_\ell$ given $\mathbf{v}_i$, for every $\ell \in \{1, \ldots, c\}$ as follows:
\begin{align*}
    P(\omega_\ell\vert \mathbf{v}_i) = \frac{P(\omega_\ell)P(\mathbf{v}_i\vert \omega_\ell)}{P(\mathbf{v}_i)}.
\end{align*}
We calculate the above posterior probabilities using the assumption of Linear Discriminant Analysis (LDA). Therefore, we obtain the posterior probabilities as 
\begin{align}
    \label{posterior in space i: LDA}
    P(\omega_\ell\vert \mathbf{v}_i) & = \frac{P(\omega_\ell)P(\mathbf{v}_i\vert \omega_\ell)}{P(\mathbf{v}_i)} \notag \\ 
    &= \frac{\exp\{-\frac{1}{2}(\mathbf{v}_i - \bm{\eta}_\ell)'\Sigma^{-1}(\mathbf{v}_i - \bm{\eta}_\ell)\}}{\sum\limits_{\ell=1}^{c} \exp\{-\frac{1}{2}(\mathbf{v}_i - \bm{\eta}_\ell)'\Sigma^{-1}(\mathbf{v}_i - \bm{\eta}_\ell)\}},
\end{align}
where $\bm{\eta}_\ell$ and $\Sigma$ are estimated from the coefficient matrix $Z_{\ell}$, defined in equation (\ref{coefficient matrix}). Each row of the $N\times q_i+c-1$ order matrix $Z_{\ell}$ is assumed to be an observation from the multivariate normal with mean $\bm{\eta}_\ell$ and common covariance matrix $\Sigma$ (standard assumption of LDA \cite{bishop2006pattern}). Therefore, the estimate of $\bm{\eta}_\ell$ is given by 
$$ \widehat{\bm{\eta}}_\ell = \frac{1}{n_\ell} \sum\limits_{k=1}^{N}\bm{z}_k I(\bm{z}_k\in \omega_{\ell}),$$ where $\bm{z}_k'$ denotes the $k$th row of the matrix $Z_\ell$ and $I$ is the indicator function defined as $I(x\in A) = 1 , \text{ if } x\in A$ and 0 otherwise. Using $\widehat{\bm{\eta}}_\ell$ we obtain the pooled estimate of $\Sigma$ (see chapter 5 of \cite{alpaydin2021machine}) as 
$$\widehat{\Sigma}  = \frac{1}{c} \sum\limits_{i=1}^{c} \frac{1}{(n_{i}-1)}\sum\limits_{k=1}^{N} I(\bm{z}_k\in \omega_{i}) (\bm{z}_k - \widehat{\bm{\eta}}_\ell)(\bm{z}_k - \widehat{\bm{\eta}}_\ell)' .$$ 
Hence, the estimated posterior probabilities of class $\omega_\ell$ given $\bm{v}_i$ in the subspace $i$ is given by 
$$\widehat{P}(\omega_\ell\vert \mathbf{v}_i) =  \frac{\exp\left\{-\frac{1}{2}(\mathbf{v}_i - \widehat{\bm{\eta}}_\ell)'\widehat{\Sigma}^{-1}(\mathbf{v}_i - \widehat{\bm{\eta}}_\ell)\right\}}{\sum\limits_{\ell=1}^{c} \exp\left\{-\frac{1}{2}(\mathbf{v}_i - \widehat{\bm{\eta}}_\ell)'\widehat{\Sigma}^{-1}(\mathbf{v}_i - \widehat{\bm{\eta}}_\ell)\right\}}.$$ 
Now, within the subspace $i$, we calculate the maximum of the these estimated posterior probabilities and the corresponding class $k$. To be precise, we define $k^{(i)} = \argmax\limits_{\ell\in \{1,\ldots, c\}} \widehat{P}(\omega_\ell\vert \mathbf{v}_i)$ with the maximum posterior probability $p^{(i)} = \max\limits_{\ell\in \{1,\ldots, c\}} \widehat{P}(\omega_\ell\vert \mathbf{v}_i)$. Finally, we assign the unknown functional observation $X^\ast$ to the class $k^{(j)}\in \{1, \ldots, c\}$, if $j=\argmax\limits_{i \in \{1,\ldots, c\}} p^{(i)}$. 
This classification part of the proposed method is implemented through the following algorithm.

\algcaption{Classification}
\begin{algorithmic}[1]
    \STATE{\textbf{begin}} \textbf{initialize} $X^\ast \gets $ test functional observation, \; $E_1^\ast,E_2^\ast,\ldots, E_c^\ast \gets$ feature extraction matrices

       \FOR{$i \gets 1$ to $c$}     
       
        \STATE{\textbf{projection:}}  $v_i=\langle X^\ast-\hat{\mu_i}, e \rangle$, \; $e\in E_i$
        \FOR{$l \gets 1$ to $c$} 
        \STATE{posterior:} compute $\widehat{P}(\omega_\ell\vert \mathbf{v}_i) = \frac{P(\omega_\ell)P(\mathbf{v}_i\vert \omega_\ell)}{P(\mathbf{v}_i)}$
        \ENDFOR
        \STATE{}maximize over classes $p^{(i)} = \max\limits_{1\leq l \leq c} \widehat{P}(\omega_\ell\vert \mathbf{v}_i)$
        \ENDFOR
        \STATE{}maximize over subspaces $j=\argmax\limits_{1 \leq i \leq c } p^{(i)}$
        \STATE{\textbf{return:}} $j$ 
        \STATE{\textbf{end}}

\end{algorithmic}

\section{Results}
\label{experiment_details}
In this section, we demonstrate the performance of our method, FCPCA, on various simulated and real data sets and compare with other nine common classification techniques. Out of these nine classification techniques three are multivariate machine learning techniques, namely, Random Forests (ranger) \cite{wright2015ranger}, a local PCA based method, CPCA, \cite{das2009efficient} and Extreme Gradient Boosting (xgboost) \cite{chen2016xgboost}), and six algorithms specifically adapted  for FDA, namely, Functional Generalized Linear Models (classif.glm) \cite{mccullagh1989binary}, Functional k-Nearest Neighbor (classif.knn) \cite{ferraty2006nonparametric}, Non-parametric Kernel classifier with asymmetric normal kernel used (classif.np) \cite{ferraty2006nonparametric}, functional classification using ML algotithms for  functional explanatory variables (classif.svm and classif.lda) \cite{ramsay2008functional,mccullagh1989binary}, and knn with dtw \cite{rakthanmanon2013addressing}. Apart from CPCA, the codes of all other eight algorithms are taken from the R packages. To perform FCPCA, we have used  B spline basis functions of order $6$ to represent the functional data. The number of functional PCs in each subspace are chosen such that at least $90\%$ of the data variability is retained. The code is developed in R (Version: 3.5.1). A detailed comparison of execution time for all the algorithms is provided in TABLEs~\ref{simulated_runtime}, \ref{runtime_UCR_data}  and  \ref{eeg_runtime}. All the experiments were carried out in the computer system with $16.0$ GB of Installed memory (RAM), $\times64$-based processor, $64$-bit Operating System (Windows 10 Pro) and Intel(R) Core(TM) processor (i5-6500 CPU @3.2GHz 3.19GHz).

\subsection{Simulation studies}
We provide ten simulation studies for depicting the performance of the proposed method. 
%The first three simulation studies are from Brownian motion, fourth one is based on a variation of Brownian motion and the fifth one is based on Gaussian process. 
For all the simulation studies, the argument of the random functions lies in $[0,1]$, that is, if $X(t)$ is a random function then $t\in [0,1]$. Also, it is assumed that the time points on which the functions are observed are equally spaced. {All the simulated data sets are plotted in Fig \ref{simulated_data}}.  %We have used B spline basis functions to represent functional data and the choices of parameters are described in the Appendix section.
For each simulation studies, we have performed a 10 fold cross validation. %for evaluating the proposed method. 
The comparison results are given in TABLE \ref{simulated_result}. The simulation details are given below:
\subsubsection{Brownian Motion with different drifts, different variances}
\label{BM with drift}
\begin{figure*}[!t]
\centering
\includegraphics[width=7 in]{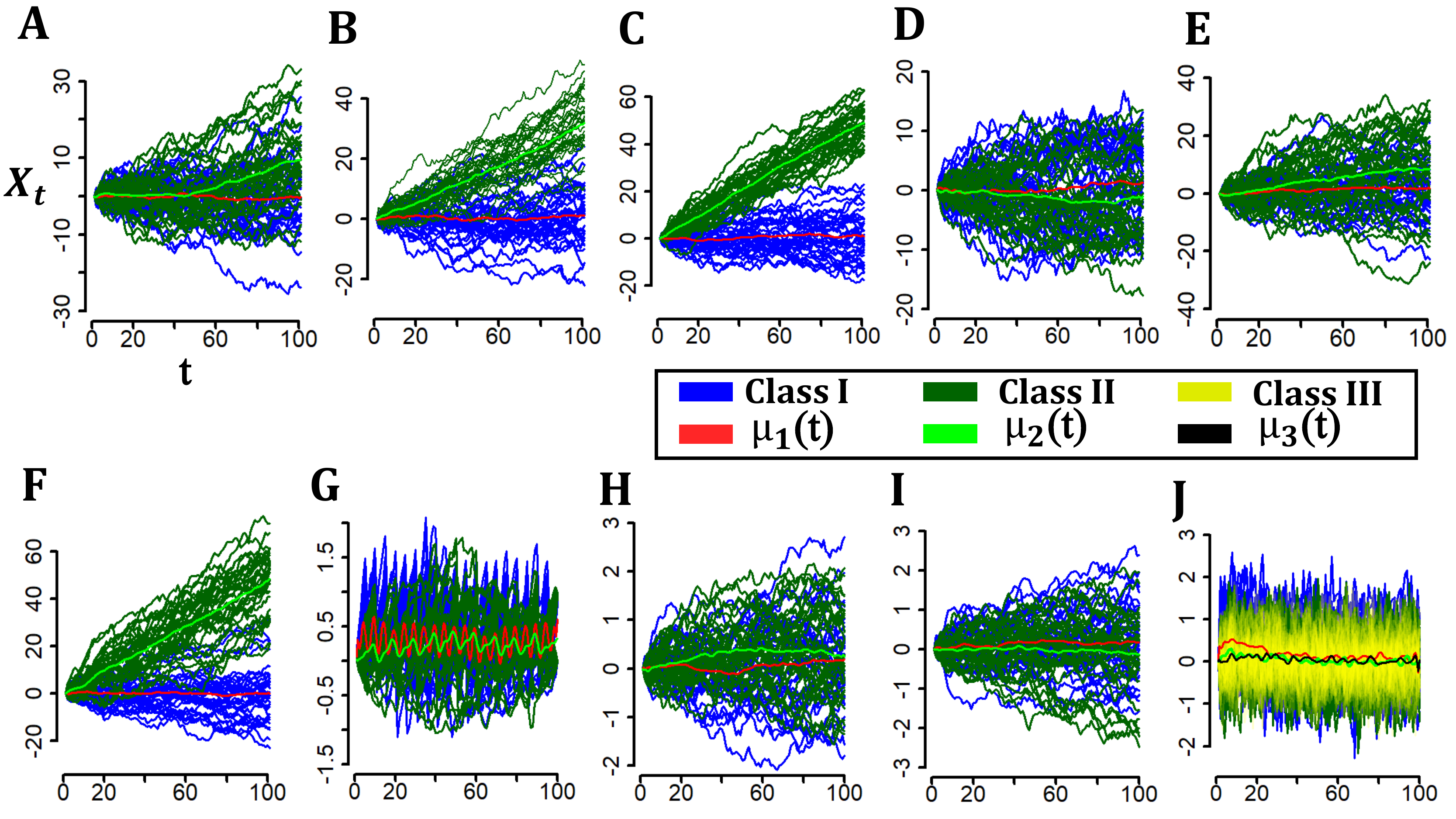}
\caption{Simulation studies: (A) BMDD1 (B) BMDD2 (C) BMDD3 (D) BMDV (E) BMDDV1 (F) BMDDV2 (G) BMCP (H) GPDM1 (I) GPDM2 (J) GP3  }
\label{simulated_data}
\end{figure*}

We have done an extensive two-class classification study on Brownian motion (BM). The class-I data sets are formed with 35 random observations from standard-BM, i.e., drift $\mu = 0$, and variance $\sigma^2 =1$. Observations of class-II are being formed with the following choices: 35 observations each from BM with $(\mu,\sigma^2) = (0.1, 1), (0.3,1), (0.5,1), (0,0.5^2), (0.1,2^2) \mbox{ and } (0.5,2^2)$, indicated as BMDD1, BMDD2, BMDD3, BMDV, BMDDV1 and BMDDV2 in the TABLE \ref{simulated_result},  respectively. So, in the first three data sets (i.e., BMDD1, BMDD2 and BMDD3) the change in drift parameter for class-II increases gradually, whereas the noise being same for both the classes ({plots of a simulated data sets for BMDD1, BMDD2 and BMDD3 are provided in Figure~\ref{simulated_data} (A), (B) and (C), respectively, along with the mean functions}). In BMDV {(depicted in Figure~\ref{simulated_data} (D))}, the noise is reduced for class-II, but the drift parameter remains the same for both the classes. And in the remaining two data sets (i.e., BMDDV1 and BMDDV2), there is a gradual and constant increase in drift and noise level for the class-II respectively. {Example of simulated data sets from BMDDV1 and BMDDV2 with their mean functions are given in Figure~\ref{simulated_data} (E) and (F).} The results show that our proposed FCPCA outperforms both machine learning and FDA based classification techniques consistently with relative percentage improvement varying at least from $1.91\%$ to at most $43.75\%$ when compared with other methods. 

%We generate $35$ functional observations from standard Brownian motion (BM) and $35$ observations from BM with a drift $\mu$. Observations from standard BM constitute class one and observations from BM with drift formed the class 2 for our classification problem.  We performed the comparison studies for $\mu = 0.1, 0.3$ and $0.5$ (data sets: BMDD1, BMDD2 and BMDD3, respectively). 

%\subsubsection{Brownian Motion with different variance}
%\label{BM with sigma}
%We generate $35$ functional observations from standard BM and $35$ observations from BM with a different variance $\sigma^2$. As in last subsection, observations from standard BM are labelled as class 1 observations and observations from BM with different variance are called class 2 observations. We have taken $\sigma^2 = 0.5^2$ for the class 2 observations (data set: BMDV). 

%\subsubsection{Brownian Motion with different drift and variance}
%\label{BM with different drift and sigma}
%We further compare our method with the other competing methods for observations generated from BM with different choices of drift ($\mu$) and variance ($\sigma^2$). We simulate $35$ functional observations from standard BM for class 1 and 35 functional observations from BM with drift $\mu$ and variance $\sigma^2$. We consider two sets of choices for $\mu$ and $\sigma^2$. We take $(\mu, \sigma^2) = (0.1, 2^2)$ and $(\mu, \sigma^2) = (0.5, 2^2)$, respectively, for two simulation studies (data sets: BMDDV1 and BMDDV2, respectively).

\subsubsection{Brownian Motion with change points}
\label{BM with change points}
{The final simulation study related to BM is done in this subsection, where the observations from class-I has a change point in its functional mean after $j$th observation and observations from class-II has a change point in its mean functional after $j'$th observation. To make this precise, we provide the details below. }
%\par 
Functional observations $X_{i}^{k}(t), \, i=1, \ldots, 30$, from class-k, $k=1,2$, follow the model  
\begin{align*}
    X_{i}^{k}(t) = \begin{cases}
    Y_{i}(t) & 1\leq i\leq 10k, \\
    \mu(t)+ Y_{i}(t) & 10k+1\leq i\leq 30,
    \end{cases}
\end{align*}
%where $Y_{i}(t),\, i=1, \ldots, 30$ are independent standard BM. 
%and the observations, $Z_{i}(t), \, 1\leq i\leq 30$, of class-II follow
%Hence, the observations of class-I have a change point at 10. Similarly, the observations, $Z_{i}(t), \, 1\leq i\leq 30$, of class-II have a change point at 20 with everything else remain the same as for class-I. That is, $Z_i(t)$, for $i=1, \ldots, 30$, follow
%\begin{align*}
%    Z_{i}(t) = \begin{cases}
%    Y_{i}(t) & 1\leq i\leq 20 \\
%    \mu(t)+ Y_{i}(t) & 21\leq i\leq 30,
%    \end{cases}
%\end{align*}
where $Y_{i}(t),\, i=1, \ldots, 30$ are independent standard BM. The mean function $\mu(t)$ is taken be $t$. Notice that the observations of class-I, class-II have a change point at 10 and at 20, respectively. This data set is indicated as BMCP in TABLE \ref{simulated_result}. {Both classes are depicted in the Figure~\ref{simulated_data} (G). The change points are not apparent in the figure due to random structure of Brownian bridge which is a dominant part of the model.} The Performance of FCPCA was superior to the other classification techniques used with relative improvement varying from $3.02\%$ to $39.98\%$.

\subsubsection{Gaussian process with different mean functions}
%Our method along with other classification methods of functional data are tested on the Gaussian process with different mean functions. 
Thirty-five observations each from class-I and class-II are generated from Gaussian process with mean function $\mu(t)$ and covariance function as $\mbox{cov}(X(t),X(s)) = \min\{s,t\}$. For class-I $\mu(t)$ is taken to be 0 and for class-II $\mu(t)$ is assigned as $t(1-t)$ and $t^2(1-t)^2$, respectively, for two separate studies (data sets indicated as GPDM1 and GPDM2 in TABLE \ref{simulated_result}, respectively). Since the mean for GPDM2 is a polynomial of degree 4 and the mean for GPDM1 is a polynomial of degree 2, the mean differences between two classes become much narrower in GPDM2 as compared to GPDM1. However, the both data sets preserve the same covariance structure for both classes. {Example data sets of GPDM1 and GPDM2, along with the mean functions, are shown in the Figure~\ref{simulated_data} (H) and (I). }  The proposed method fared well when compared to other techniques and relative improvement was in the range of $0.36\%-37.15\%$.

%Another independent set of 35 observations, constituting class-II observations, are generated from Gaussian process with mean function $\mu(t)$ and the same covariance function as above. 
%These observations constitute the class-II for our classification problem. 
%We have done two separate studies for  (a) $\mu(t) = t(1-t)$, and (b) $\mu(t) = t^2(1-t)^2$ 

\subsubsection{3-class problem using Gaussian process}
\label{GP with sine function}
In this subsection, we have compared our method with the others for a three class classification problem. This particular simulation study is structurally different than others done above, in the sense that, there are repeated observations in each class. For each class we add a sinusoidal component with random coefficients (see \cite{aguilera2020multi} for details). {A plot of sample data set with three mean functions are provided in Figure~\ref{simulated_data} (J). The complete description of the model is as follows:}
Let $X_{i}^k$ denote the $i$th functional observation from class $k$, for $k=1,2,3$. $X_{i}^k$ follows the following model:
\begin{align*}
    X_{i}^k = m^k(t) + a_i \sin(\pi t) + \epsilon_{i}^k(t),
\end{align*}
where $i=1, \ldots, 40$, $k=1,2,3$, $m^k(t)=t^{\frac{k}{5}}(1-t)^{(6-{\frac{k}{5}})}$, $a_i\sim N(\mu_i,0.02^2)$, $\mu_i \sim \mbox{Unif}(0,0.05)$ and $\epsilon_i^k(t)\sim \mbox{GP}$ with 0 mean and covariance function to be identity, for all $i$ and $k$ (data set: GP3). Here, GP stands for Gaussian process. 
%This model is taken from the paper by Aguilera et al. \cite{aguilera2020multi}. 
Although our proposed method is not manufactured for repeated measurements, as done in \cite{aguilera2020multi}, still 
the performance of FCPCA was better than the other methods, compared in this article, for this simulated data and relative improvement varied in between $2.6\%-54.11\%$.% It is important to mention here that the performance of our method is worse than that of \cite{aguilera2020multi} for this data set and it will be our future project to improve our method for repeated measurements. 

\begin{table}
\renewcommand{\arraystretch}{1.3}
  \caption{Comparison of FCPCA with other techniques of classification for functional data using  10 simulated data sets. Mean classification accuracies along with corresponding standard deviations (in parenthesis) are reported. Maximum accuracy for a data set is identified in bold.  }
  \label{simulated_result}
  \centering
  \resizebox{\columnwidth}{!}{\begin{tabular}{llllll}
    \toprule
    Data sets & \textbf{FCPCA} & \textbf{classif.glm} & \textbf{classif.knn} & \textbf{classif.np}  & \textbf{classif.svm}    \\\cmidrule(r){1-6}
    BMDD1 & \textbf{0.752} (0.17)& 0.723 (0.17) & 0.664 (0.18) & 0.680 (0.19) & 0.599 (0.19) \\
    BMDD2 & \textbf{0.900} (0.12) & 0.869 (0.13) & 0.898 (0.12) & 0.866 (0.13) & 0.864 (0.14)\\
    BMDD3 & \textbf{0.970} (0.06) & 0.939 (0.10) & 0.959 (0.07) & 0.927 (0.09) & 0.941 (0.09)\\
    BMDV & \textbf{0.609} (0.19) & 0.556 (0.19) & 0.503 (0.17) & 0.528 (0.15) & 0.385 (0.16) \\
     BMDDV1 & \textbf{0.679} (0.18) & 0.632 (0.19) & 0.640 (0.18) & 0.666 (0.18) & 0.613 (0.18) \\
     BMDDV2 & \textbf{0.938} (0.10) & 0.913 (0.11) & 0.891 (0.12) & 0.912(0.11) & 0.928 (0.10) \\
     BMCP & \textbf{0.828} (0.14) & 0.637 (0.18) & 0.738 (0.17) & 0.736 (0.18) & 0.497 (0.21) \\
      GPDM1 & \textbf{0.638} (0.18) & 0.587 (0.19) & 0.532 (0.19) & 0.548 (0.19) & 0.401 (0.17)\\
      GPDM2 & \textbf{0.553} (0.19) & 0.538 (0.20) & 0.442 (0.19) & 0.399 (0.18) & 0.551 (0.17) \\
      GP3 & \textbf{0.632} (0.13) & 0.603 (0.11) & 0.394 (0.12) & 0.365 (0.11) & 0.558 (0.13) \\
    \cmidrule(r){2-6}
     & \textbf{classif.lda} & \textbf{knn dtw} & \textbf{ranger} & \textbf{CPCA}  & \textbf{xgboost} \\\cmidrule(r){2-6}
   BMDD1 & 0.726 (0.18)& 0.423 (0.19) & 0.675 (0.19) & 0.738 (0.17) & 0.575 (0.20) \\
   BMDD2 & 0.863 (0.13) & 0.592 (0.19) & 0.876 (0.13) & 0.880 (0.13) & 0.899 (0.12) \\
   BMDD3 & 0.959 (0.07) & 0.598 (0.18) & 0.944 (0.09) & 0.946 (0.09) & 0.912 (0.10) \\
   BMDV & 0.593 (0.20) & 0.607 (0.20) & 0.552 (0.20) & 0.544 (0.17) & 0.553 (0.19) \\
   BMDDV1 & 0.641 (0.18) & 0.412 (0.19) & 0.642 (0.18) & 0.605 (0.20) & 0.613 (0.19)\\
   BMDDV2 & 0.926 (0.10) & 0.626 (0.19) & 0.897 (0.11) & 0.911 (0.11) & 0.910 (0.11)\\
   BMCP & 0.681 (0.18) & 0.765 (0.16) & 0.803 (0.16) & 0.777 (0.16) & 0.632 (0.18)\\
   GPDM1 & 0.617 (0.20) & 0.541(0.19) & 0.603 (0.19) & 0.617 (0.19) & 0.598 (0.19) \\
   GPDM2 & 0.500 (0.19) & 0.389 (0.20) & 0.396 (0.19) & 0.462 (0.19) & 0.426 (0.20)\\
   GP3 & 0.615 (0.13) & 0.290 (0.12) & 0.402 (0.13) & 0.446 (0.14) & 0.392 (0.13) \\
    \bottomrule
  \end{tabular}}
\end{table}

\subsection{Benchmarked DataSets}

In order to obtain the effectiveness of FCPCA, we selected  10 time series data sets, with various application types such as ECG, sensor, food spectrograph, or image outline data, from the popular UEA \& UCR time series classification repository \cite{bagnall2018uea}.  Instead of considering a single test/train split, we run 100 resampling folds on each of the selected data sets as suggested in the article by Bagnall et al. \cite{bagnall2017great}. The details of the data sets with split information are described in TABLE~\ref{UCR_data_description} and the results are shown in TABLE \ref{UEA_UCR_result}.
%\footnote{Performance using Test/Train datasplit also gave similar results but we have used validation techniques to maintain uniformity of results}. 
As seen from the results, FCPCA fares favorably when compared to other methods and has the best overall performance in most of the datasets (6 out of 10) with relative improvement varying from $2.13\%-65.34\%$.

\begin{table}
  \caption{Details of time series data sets from the popular UEA \& UCR time series classification repository  }
  \label{UCR_data_description}
  \centering
  \resizebox{\columnwidth}{!}{\begin{tabular}{lllll}
    \toprule
    Name & \textbf{Obs. (Train + Test)} & \textbf{Length} & \textbf{Classes} &  \textbf{Type}    \\\hline 
    Beef & 60 (30 + 30)& 470 & 5  & SPECTRO \\
    CBF & 930 (30 + 900) & 128 & 3  & SIMULATED \\
    Car & 120 (60 + 60) & 577 & 4 & SENSOR\\
    ECGFiveDays &884 (23 + 861) & 136 & 2 & ECG \\
    Fish & 350 (175 + 175) & 463 & 7 & IMAGE \\
    Ham & 214 (109 + 105) & 431 & 2 & SPECTRO\\
    Mote Strain & 1272 (20 + 1252) & 84 & 2 & SENSOR \\
    Strawberry & 983 (613 + 370) & 235 & 2 & SPECTRO\\
    SyntheticControl & 600 (300 + 300)& 60 & 6 & SIMULATED\\
    TwoLeadECG & 1162 (23 + 1139) & 82 & 2 & ECG \\
    \bottomrule
  \end{tabular}}
\end{table}

\begin{table}
  \caption{Comparison of FCPCA with other techniques of classification for functional data using  10  data sets from the UEA \& UCR time series classification repository. Mean classification accuracies along with corresponding standard deviations (in parenthesis) are reported. Maximum accuracy for a data set is identified in bold.  }
  \label{UEA_UCR_result}
  \centering
  \resizebox{\columnwidth}{!}{\begin{tabular}{llllll}
    \toprule
    Data sets & \textbf{FCPCA} & \textbf{classif.glm} & \textbf{classif.knn} & \textbf{classif.np}  & \textbf{classif.svm}    \\\hline 
    Beef & \textbf{0.782} (0.08)& 0.498 (0.09) & 0.363 (0.09) & 0.271 (0.06) & 0.466 (0.09) \\
    CBF & 0.876 (0.04) & 0.857 (0.05) & 0.796 (0.06) & 0.546 (0.12) & 0.849 (0.04) \\
    Car & \textbf{0.776} (0.06) & 0.612 (0.06) & 0.685 (0.06) & 0.466 (0.09) & 0.635 (0.06) \\
    ECGFiveDays & \textbf{0.908} (0.05) & 0.805 (0.06) & 0.710 (0.09) & 0.550 (0.09) & 0.724 (0.11)\\
    Fish & \textbf{0.797} (0.03) & 0.492 (0.03) & 0.761 (0.03) & 0.278 (0.09) & 0.508 (0.04)\\
    Ham &\textbf{0.821}  (0.04) & 0.633 (0.04) & 0.735 (0.05) & 0.599 (0.10) & 0.590 (0.08) \\
    Mote Strain & 0.856 (0.03) & 0.826 (0.05) & 0.834 (0.05) & 0.605 (0.13) & 0.759 (0.13)\\
    Strawberry & 0.951 (0.01) & 0.815 (0.02) & 0.937 (0.01) & 0.644 (0.02) & 0.867 (0.02)\\
    SyntheticControl & 0.952 (0.01) & 0.858 (0.02) & 0.879 (0.02) & 0.477 (0.09) & 0.950 (0.01)\\
    TwoLeadECG & \textbf{0.881} (0.05) & 0.755 (0.05) & 0.624 (0.06) & 0.508 (0.02) & 0.517 (0.10) \\\cmidrule(r){2-6}
     & \textbf{classif.lda} & \textbf{knn dtw} & \textbf{ranger} & \textbf{CPCA}  & \textbf{xgboost} \\\cmidrule(r){2-6}
   Beef & 0.633 (0.08)& 0.602 (0.09) & 0.552 (0.10) & 0.476 (0.11) & 0.466 (0.10) \\
   CBF & \textbf{0.903} (0.02) & 0.351 (0.02) & 0.875 (0.05) & 0.731 (0.10) & 0.732 (0.05) \\
   Car & 0.671 (0.05) & 0.621 (0.05) & 0.696 (0.05) & 0.493 (0.06) & 0.570 (0.07)\\
   ECGFiveDays & 0.813 (0.06) & 0.791 (0.04) & 0.809 (0.06) & 0.788 (0.09) & 0.743 (0.07) \\
   Fish & 0.531 (0.03) & 0.759 (0.03) & 0.780 (0.03) & 0.501 (0.04) & 0.626 (0.04) \\
   Ham & 0.665 (0.04) & 0.778 (0.03) & \textbf{0.821} (0.04) & 0.762 (0.05) & 0.710 (0.05)\\
   Mote Strain & 0.846 (0.04) & \textbf{0.905} (0.03) & 0.854 (0.04) & 0.850 (0.03) & 0.773 (0.06) \\
   Strawberry & 0.772 (0.02) & 0.959 (0.01) & \textbf{0.966} (0.01) & 0.684 (0.03) &0.941 (0.01)\\
   SyntheticControl & 0.901 (0.01) & 0.386 (0.02) & \textbf{0.964} (0.01) & 0.762 (0.03) & 0.749 (0.03)\\
   TwoLeadECG  & 0.756 (0.04) & 0.739 (0.04) & 0.787 (0.05) & 0.689 (0.08) & 0.728 (0.06)\\
    \bottomrule
  \end{tabular}}
\end{table}
\begin{figure*}[!t]
\centering
\includegraphics[width=6 in]{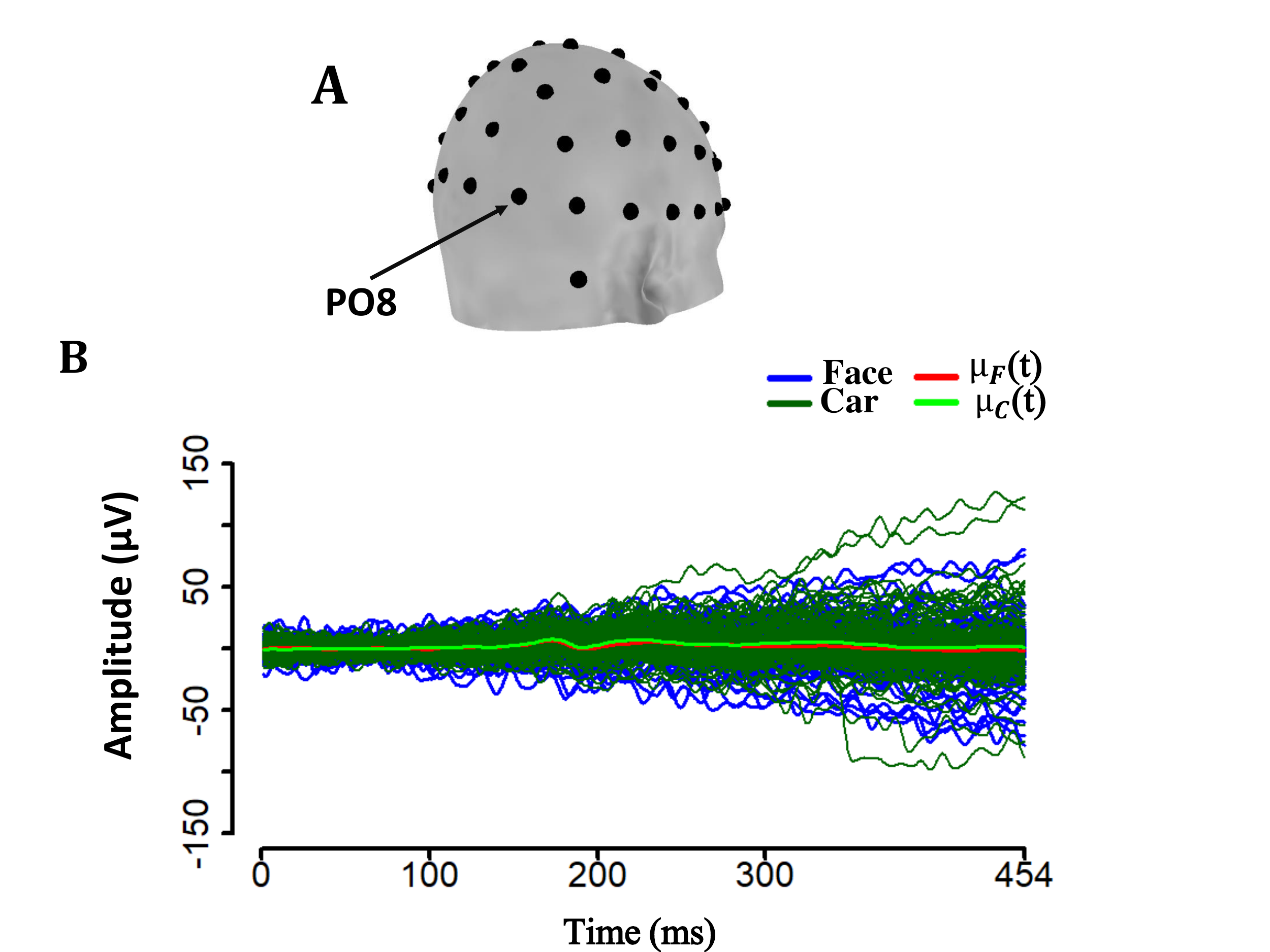}
\caption{ (A) Electrode positions on human scalp: The electrode "PO8" is identified. Electrode positions are mapped using sLORETA software \cite{pascual2002standardized}. (B) EEG data of the "PO8" electrode: Two classes (Face and Car) each having 125 samples. Corresponding mean functions ($\mu_i(t)
\;i=F,C$ where $F$ and $C$ denote Face and Car respectively) are also plotted. }
\label{eeg_face_car}
\end{figure*}

\subsection{Neural (EEG) dataset}
The final dataset we used involved previously published EEG signals which involved a face/car identification task \cite{saha2020our}. Since previous literature \cite{rossion2003early} shows differential neural activity in parieto-occipital electrodes, we performed the  classification task between face and car for the PO8 electrode (shown in Figure \ref{eeg_face_car} (A)). This classification task consists two classes each having $125$ EEG signals and $454$ time points. EEG signals, along with their mean functions, corresponding to PO8 for face and car are depicted in Figure~\ref{eeg_face_car} (B). Total $17$ subjects participated in that experiment. We have done a $10$ fold cross validation for subject wise classification and the mean classification accuracy across subjects are mentioned in TABLE~\ref{eeg_result}. Neural datasets are typically noisy and hard to classify but as seen from the results, our method performs well compared to others even in a hard classification task.

\begin{table}
  \caption{Comparison of FCPCA with other techniques of classification for EEG data. Mean classification accuracy along with corresponding standard deviations (in parenthesis)  are reported. Maximum accuracy for a data set is identified in bold.  }
  \label{eeg_result}
  \centering
   \resizebox{\columnwidth}{!}{\begin{tabular}{lllll}
    \toprule
   \textbf{FCPCA} & \textbf{classif.glm} & \textbf{classif.knn} & \textbf{classif.np}  & \textbf{classif.svm}    \\\cmidrule(r){1-5}
     \textbf{0.605} (0.06)& 0.545 (0.06) & 0.559 (0.06) & 0.525 (0.05) & 0.522 (0.07) \\
    
    \cmidrule(r){1-5}
     \textbf{classif.lda} & \textbf{knn dtw} & \textbf{ranger} & \textbf{CPCA}  & \textbf{xgboost} \\\cmidrule(r){1-5}
    0.557 (0.06)& 0.512 (0.04) & 0.576 (0.07) & 0.565 (0.06) & 0.539 (0.04) \\
    
    \bottomrule
  \end{tabular}}
\end{table}

\section{Discussion and Conclusion}
We present here a novel classification framework using functional data analysis which can be used for classifying high dimensional data successfully. Our proposed method performs well when compared to both functional data analysis techniques as well as state-of-the-art machine learning techniques using variety of data sets. 
Focusing on the classification accuracy from the synthetic datasets, we observe that when the drift of the brownian motion is increased, the classification increases as expected while the performance reduces with increase in the variance using all the classifiers. However, using data generated from Gaussian Process, the advantage of using FCPCA becomes more prominent. The mean difference between the two classes reduces more in the data set  GPDM2 compared to GPDM1, however, apart from FCPCA, functional SVM and functional GLM classifier, almost all classifiers give below chance performance. Thus for time series data with narrow mean difference, classification using FDA techniques especially using FCPCA  gives greater advantage. 

Our technique is particularly suited for hard classification tasks involving noisy signals as demonstrated by its superior performance using synthetic data with change point and real data using  EEG signals. In both these data sets, FCPCA outperforms all other techniques by a significant margin. The improvement produced by FCPCA can possibly be attributed to the non-linear nature of the proposed functional feature extraction step which aids in capturing the underlying data manifold by projecting in multiple subspaces. It is noteworthy to mention here that our proposed framework has the option of further reducing the data in the projected classwise FPCA subspace using any linear data transformation techniques of choice (\cite{das2009efficient}) although the current implementation does not use any further data reduction method after functional classwise PCA features are obtained. The superior performance of FCPCA over other techniques in both simulated and real datasets with various degrees of noise demonstrates that FCPCA acts as a powerful data reduction and classification framework.

The computational time of FCPCA (TABLEs~\ref{simulated_runtime}, \ref{runtime_UCR_data} and \ref{eeg_runtime}) shows that although it does not give the least computation time, but in most cases, it gives average run-time and the longer run-time can be attributed to the piecewise linear nature of the method. However, the efficacy of the method can also be attributed to the very nature of piecewise linear feature extraction which makes it suitable for use in classifying highly overlapped classes. One major drawback of the method is that it does not scale favorably to number of classes since number of functional projection subspace depends on number of classes. The performance of the method might be affected if the number of observations for different classes vary greatly. Unbalanced training data can give rise to inaccurate estimates of covariance functions for different classes leading to unstable results. Another drawback of our and any functional data analysis based method is the inability of projecting the feature extraction matrix back the original data space in order to identify relevant spatiotemporal regions carrying discriminatory information similar to Eigenface (see \cite{turk1991eigenfaces}) or Fisherface (refer to \cite{belhumeur1996eigenfaces}). While converting data to the functional form, the spatio-temporal information in the original space is lost and hence it cannot be reconstructed back. However, this step of converting data points into a functional form gives the added advantage of removing small sample size problem as the functional data is expressed in terms of a few functional coefficients (typically Fourier or B-Spline coefficients) thus reducing the data dimension drastically and making the functional space non-sparse.

%\subsection{Broader Impact}
%\section{Conclusion}
Functional data analysis has garnered recent attention for its use in high dimensional data analysis and classification. The concept of functional data analysis makes it ideal for applications suffering from small sample size problem. The method proposed in the paper offers a novel data reduction and classification framework using functional data analysis which can act as an efficient supervised classification tool, especially for high dimensional time-series data. Efficacy of the proposed method for easy and hard classification tasks as well as its performance using data in various fields from biology to economics alludes to its potential as a viable alternative in supervised learning techniques. 

\begin{table}
  \caption{Comparison of the Run time of FCPCA with other algorithms  using  10  simulated data sets. Run time for a single train and test run is recorded in second. }
  \label{simulated_runtime}
  \centering
  \resizebox{\columnwidth}{!}{\begin{tabular}{llllll}
    \toprule
    Data sets & \textbf{FCPCA} & \textbf{classif.glm} & \textbf{classif.knn} & \textbf{classif.np}  & \textbf{classif.svm}    \\\cmidrule(r){1-6}
    BMDD1 & 0.17& 0.33 & 0.37 & 0.36 & 0.35 \\
    BMDD2 & 0.26 & 0.17 & 0.56 &  0.36 &  0.13\\
    BMDD3 & 0.31 & 0.18 & 0.32 & 0.17 & 0.36\\
    BMDV & 0.19 & 0.16 & 0.36 & 0.17 & 0.37 \\
     BMDDV1 & 0.19 & 0.16 & 0.31 & 0.19 & 0.33 \\
     BMDDV2 & 0.18 & 0.14 & 0.39 & 0.36 & 0.35 \\
     BMCP & 0.32 & 0.11 & 0.31 & 0.22 & 0.39 \\
      GPDM1 & 0.18 & 0.16 &  0.33 & 0.36 & 0.15\\
      GPDM2 & 0.18 & 0.15 & 0.30 & 0.37 & 0.15 \\
      GP3 & 0.18 & 0.16 & 0.68 & 0.61 & 0.09 \\
    \cmidrule(r){2-6}
     & \textbf{classif.lda} & \textbf{knn dtw} & \textbf{ranger} & \textbf{CPCA}  & \textbf{xgboost} \\\cmidrule(r){2-6}
   BMDD1 & 0.13 & 2.29 & 0.21 & 0.18 & 0.09 \\
   BMDD2 & 0.13 & 1.89 &0.20 & 0.19 & 0.10 \\
   BMDD3 & 0.07 & 1.86 & 0.22 & 0.18 & 0.10 \\
   BMDV & 0.14 & 1.85 & 0.10 & 0.18 & 0.10 \\
   BMDDV1 & 0.14 & 0.50 & 0.22 & 0.23 & 0.09\\
   BMDDV2 &  0.19 & 2.03 & 0.18 &0.19 & 0.08\\
   BMCP & 0.10 &1.73 & 0.13 & 0.19 & 0.08 \\
   GPDM1 & 0.08 & 2.03 & 0.08 & 0.19 & 0.09 \\
   GPDM2 & 0.09 & 2.05 & 0.08 & 0.19 &  0.10\\
   GP3 &  0.09 & 6.72 & 0.11 & 0.21 & 0.11 \\
    \bottomrule
  \end{tabular}}
\end{table}

\begin{table}
  \caption{Comparison of the Run time of FCPCA with other algorithms  using  10  data sets from the UEA \& UCR time series classification repository. Run time for a single train and test run is recorded in second.   }
  \label{runtime_UCR_data}
  \centering
  \resizebox{\columnwidth}{!}{\begin{tabular}{llllll}
    \toprule
    Data sets & \textbf{FCPCA} & \textbf{classif.glm} & \textbf{classif.knn} & \textbf{classif.np}  & \textbf{classif.svm}    \\\hline 
    Beef & 0.94& 0.50 & 0.93 & 0.33 & 0.75 \\
    CBF & 1.06 & 0.50 & 4.90 & 1.64 & 0.64 \\
    Car & 0.76 & 0.43 & 1.95 & 0.75 & 0.67 \\
    ECGFiveDays & 0.42 & 0.22 & 3.70 &1.43 & 0.68\\
    Fish & 2.09 & 0.84 & 8.23 & 3.44 & 0.95\\
    Ham & 0.41  & 0.25 & 3.67 & 1.41 & 0.78 \\
    Mote Strain & 0.60 & 0.26 & 4.92 & 1.61 & 0.67\\
    Strawberry & 0.56 & 0.29 & 20.4 & 19.6  & 0.95\\
    SyntheticControl & 1.82 & 0.70 & 5.10 & 4.58 & 1.07\\
    TwoLeadECG & 0.50 & 0.20 & 2.04 & 1.70 & 0.61 \\\cmidrule(r){2-6}
     & \textbf{classif.lda} & \textbf{knn dtw} & \textbf{ranger} & \textbf{CPCA}  & \textbf{xgboost} \\\cmidrule(r){2-6}
   Beef & 0.13& 39.14 & 0.11 & 0.55 & 0.11 \\
   CBF & 0.11 & 143.07 & 0.11 & 0.36 & 0.07 \\
   Car & 0.13 & 210.72 & 0.17 & 0.31 & 0.11\\
   ECGFiveDays & 0.13 & 135.89 & 0.11 & 0.28 & 0.08 \\
   Fish & 0.14 & 1411.58 & 0.49 & 0.45 & 0.17 \\
   Ham &0.09 & 488.68 & 0.17 & 0.25 &0.09\\
   Mote Strain & 0.14 & 94.64 & 0.11 & 0.39 & 0.10 \\
   Strawberry & 0.2 & 2688.64 & 0.57 & 0.31 &0.13\\
   SyntheticControl & 0.11 & 410.30 &0.51 & 0.45 & 0.11\\
   TwoLeadECG  & 0.17 & 120.04 & 0.31 & 0.37 & 0.06\\
    \bottomrule
  \end{tabular}}
\end{table}
\begin{table}
  \caption{Comparison of the Run time of FCPCA with other algorithms  for the EEG data set. Run time for a single train and test run is recorded in second.  }
  \label{eeg_runtime}
  \centering
  \resizebox{\columnwidth}{!}{\begin{tabular}{lllll}
    \toprule
   \textbf{FCPCA} & \textbf{classif.glm} & \textbf{classif.knn} & \textbf{classif.np}  & \textbf{classif.svm}    \\\cmidrule(r){1-5}
     0.37 & 0.22 & 1.38 & 1.21 & 0.30 \\
    
    \cmidrule(r){1-5}
     \textbf{classif.lda} & \textbf{knn dtw} & \textbf{ranger} & \textbf{CPCA}  & \textbf{xgboost} \\\cmidrule(r){1-5}
    0.31& 95.60 & 0.32 & 0.14 & 0.13 \\
   
    \bottomrule
  \end{tabular}}
\end{table}

\ifCLASSOPTIONcompsoc
  \section*{Acknowledgments}
\else
  \section*{Acknowledgment}
\fi

A. Chatterjee is supported by an INSPIRE fellowship from the Department of Science and Technology (DST), Government of India.

\ifCLASSOPTIONcaptionsoff
  \newpage
\fi



%\bibliographystyle{IEEEtran}
%\begin{thebibliography}{1}


%\end{thebibliography}

\bibliography{ref.bib}
\bibliographystyle{IEEEtran}


\begin{IEEEbiography}[{\includegraphics[width=1in,height=1.25in,clip,keepaspectratio]{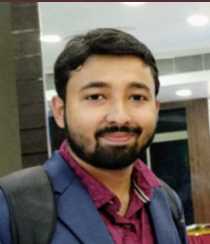}}]{Avishek Chatterjee}

Avishek Chatterjee received the M.Sc. degree in Mathematics and Computing from IIT (ISM) Dhanbad, India in 2016.
He is currently a PhD scholar of IISER Kolkata, Mohanpur, India.  His current research interests include computational neuroscience, neural correlates of decision making process, pattern recognition and application of functional data analysis. 
\end{IEEEbiography}

\begin{IEEEbiography}[{\includegraphics[width=1in,height=1.25in,clip,keepaspectratio]{author2.pdf}}]{Satyaki Mazumder}
Satyaki Mazumder received the M.Sc. degree in Statistics from IIT Kanpur, India in 2007, and the Ph.D. degree in Statistics from the University of Texas at Dallas, Dallas, in 2010.
He is currently an assistant professor of IISER Kolkata, Mohanpur, India.  His  current research interests include Bayesian modeling, application of functional data analysis, behavioural modeling and statistical modeling of EEG signals.
\end{IEEEbiography}

\begin{IEEEbiography}[{\includegraphics[width=1in,height=1.25in,clip,keepaspectratio]{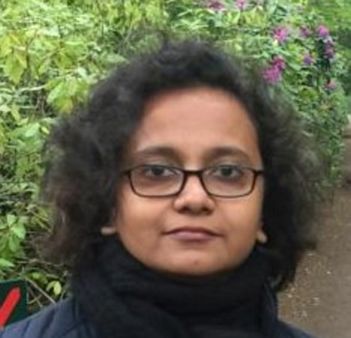}}]{Koel Das}
Koel Das received the M.S. degree in electrical engineering from Wright State University, Dayton, OH, in 2003, and the Ph.D. degree in electrical engineering and computer science from the University of California, Irvine, in 2007.
She is currently an associate professor of IISER Kolkata, Mohanpur, India.  Her  current research interests include brain–machine interfaces,pattern recognition, image processing, and geometric processing.
\end{IEEEbiography}


%\newpage



%\vfill


\end{document}